\documentclass[letterpaper]{article} 
\usepackage{aaai25}  
\usepackage{times}  
\usepackage{helvet}  
\usepackage{courier}  
\usepackage[hyphens]{url}  
\usepackage{graphicx} 
\usepackage{natbib}  
\usepackage{caption} 
\usepackage{tabularx}
\usepackage{booktabs}
\usepackage{lipsum}

\usepackage{algorithm}
\usepackage{algorithmic}
\usepackage{amsmath}

\usepackage{newfloat}
\usepackage{listings}
\usepackage{multirow}
\usepackage[most]{tcolorbox}

\DeclareCaptionStyle{ruled}{labelfont=normalfont,labelsep=colon,strut=off} 
\floatstyle{ruled}
\newfloat{listing}{tb}{lst}{}
\floatname{listing}{Listing}
\title{LLM-TA: An LLM-Enhanced Thematic Analysis Pipeline for Transcripts from Parents of Children with Congenital Heart Disease}
\author{
    Muhammad Zain Raza\textsuperscript{\rm 1}\equalcontrib,
    Jiawei Xu\textsuperscript{\rm 1}\equalcontrib,
    Terence Lim\textsuperscript{\rm 2,3},
    Lily Boddy\textsuperscript{\rm 2},
    Carlos M. Mery\textsuperscript{\rm 4,5},
    Andrew Well\textsuperscript{\rm 6,7},
    Ying Ding\textsuperscript{\rm 1,7}
}
\affiliations{
    \textsuperscript{\rm 1}School of Information, UT Austin\\
    \textsuperscript{\rm 2}College of Natural Sciences, UT Austin\\
    \textsuperscript{\rm 3}Graphen, Inc.\\
    \textsuperscript{\rm 4}Division of Pediatric Cardiac Surgery, Vanderbilt University Medical Center\\
    \textsuperscript{\rm 5}Pediatric Heart Institute at Monroe Carell Jr. Children’s Hospital at Vanderbilt\\
    \textsuperscript{\rm 6}Texas Center for Pediatric and Congenital Heart Disease\\
    \textsuperscript{\rm 7}Dell Medical School, UT Austin\\
    raza.zain08@austin.utexas.edu,
    \{jiaweixu,terence.lim\}@utexas.edu,
    lilyboddy212@gmail.com,
    carlos.mery@vumc.org,
    Andrew.Well@austin.utexas.edu,
    ying.ding@ischool.utexas.edu
}

\begin{document}

\maketitle

\begin{abstract}
Thematic Analysis (TA) is a fundamental method in healthcare research for analyzing transcript data, but it is resource-intensive and difficult to scale for large, complex datasets. This study investigates the potential of large language models (LLMs) to augment the inductive TA process in high-stakes healthcare settings. Focusing on interview transcripts from parents of children with Anomalous Aortic Origin of a Coronary Artery (AAOCA)—a rare congenital heart disease—we propose an \textbf{LLM}-Enhanced \textbf{T}hematic \textbf{A}nalysis (\textbf{LLM-TA}) pipeline. Our pipeline integrates an affordable state-of-the-art LLM (GPT-4o mini), LangChain, and prompt engineering with chunking techniques to analyze nine detailed transcripts following the inductive TA framework. We evaluate the LLM-generated themes against human-generated results using thematic similarity metrics, LLM-assisted assessments, and expert reviews. Results demonstrate that our pipeline outperforms existing LLM-assisted TA methods significantly. While the pipeline alone has not yet reached human-level quality in inductive TA, it shows great potential to improve scalability, efficiency, and accuracy while reducing analyst workload when working collaboratively with domain experts. We provide practical recommendations for incorporating LLMs into high-stakes TA workflows and emphasize the importance of close collaboration with domain experts to address challenges related to real-world applicability and dataset complexity. \url{https://github.com/jiaweixu98/LLM-TA}

\end{abstract}

%

\begin{figure*}[!ht]
    \centering
    \includegraphics[width=.8\linewidth]{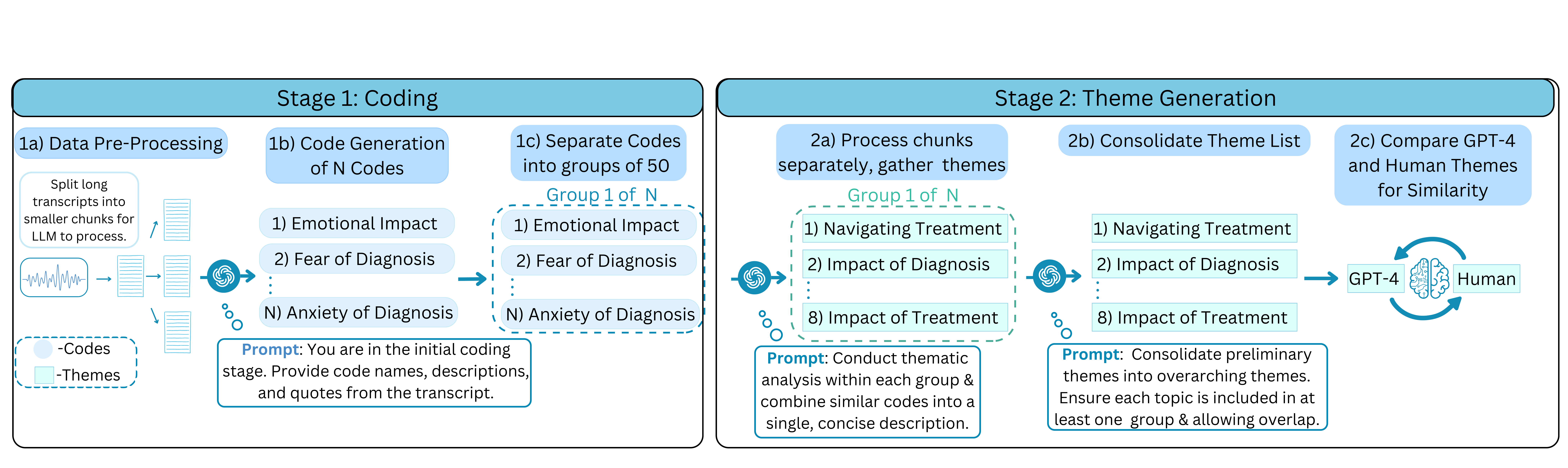}
    \caption{Pipeline for LLM-enhanced thematic analysis (LLM-TA) of transcripts from parents of children with AAOCA.}
    \label{fig:1}
\end{figure*}
\begin{figure*}[!ht]
    \centering
    \includegraphics[width=.8\linewidth]{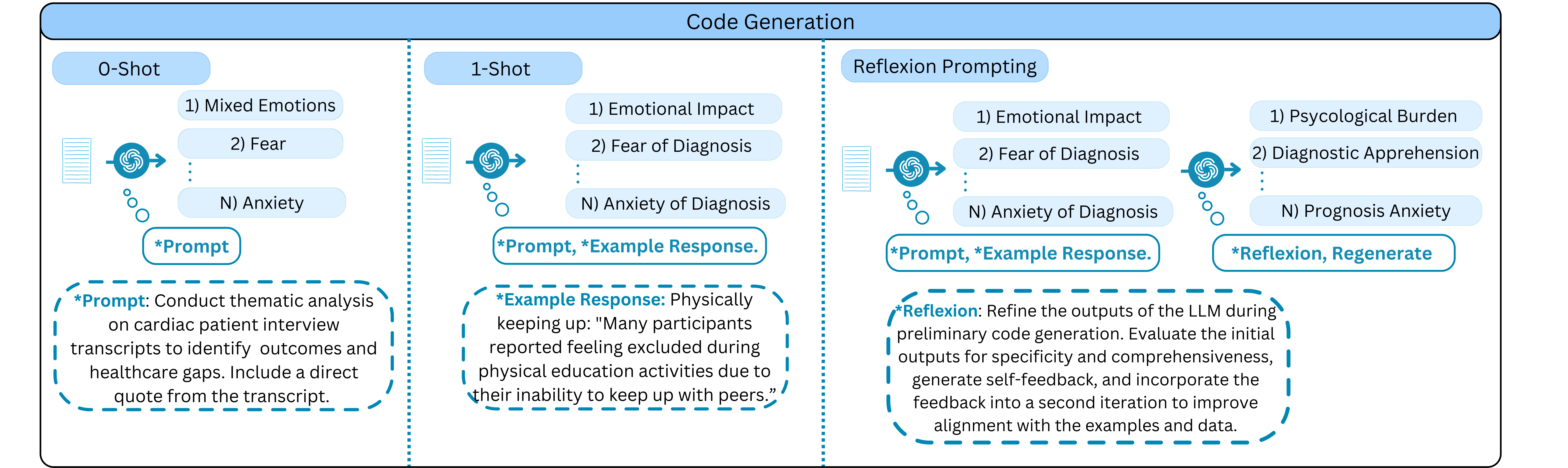}
    \caption{Prompting strategies in the LLM-TA code generation stage. These strategies are also used for theme generation.}
    \label{fig:2}
\end{figure*}

\section{Introduction}

Thematic Analysis (TA) is a widely employed qualitative research method, particularly prevalent in healthcare and related fields such as psychology~\cite{braun2006using}, heart disease research~\cite{mery2023examining}, sport and exercise studies~\cite{braun2016using}, gender identity exploration~\cite{bradford2020creating}, and anorexia research~\cite{tierney2010living}. TA facilitates an inductive examination of participant perspectives, enabling the identification of unanticipated themes, patterns, and insights in textual datasets such as patient interview transcripts~\cite{braun2006using, saldana2011fundamentals}. The seminal work by Braun and Clarke~\cite{braun2006using} defines a six-phase process for inductive TA: (1) familiarization with the data, (2) generation of initial codes, (3) theme identification, (4) theme review, (5) theme definition and naming, and (6) report production. This iterative and reflective process involves continuous movement between these phases to ensure a thorough analysis~\cite{nowell2017thematic}.

Inductive TA does not rely on pre-determined codes or themes; instead, analysts derive codes directly from the data without imposing external frameworks. However, this flexibility, while valuable for capturing nuanced insights, can lead to inconsistencies and challenges in coherence~\cite{holloway2003status}. Furthermore, the approach is resource-intensive, time-consuming, and does not scale effectively for large datasets. While small teams can manage hundreds of observations, analyzing thousands of data points introduces unique challenges related to consistency and resource allocation~\cite{katz2024thematic}. Researchers have incorporated machine learning to assist humans during the data annotation process, such as by learning patterns in real-time as user annotated the data~\cite{gebreegziabher2023patat} or leveraging rationale extraction models to generate theme recommendations~\cite{overney2024sensemate}. Some studies have explored automating the themes identification by using topic modeling techniques~\cite{leeson-2019,guetterman-2018}. However, these methods are primarily statistical, only considering the prevalence of keywords in the corpus and do not capture the nuances of human researchers' results~\cite{parfenova-etal-2024-automating}.

To address these limitations, several studies have explored the potential of large language models (LLMs) to automate inductive TA tasks. These efforts include evaluating LLMs' alignment with human annotations in diverse contexts such as video game players and data science educators~\cite{de2024performing}, psychiatric patient-clinician interactions~\cite{mathis2024inductive}, pandemic-era team feedback~\cite{katz2024thematic}, barriers to arthroplasty~\cite{mannstadt2024novel}, vaccine rhetoric on Twitter~\cite{deiner2024large}, media coverage of a controversial financial scandal~\cite{khan2024automatingthematicanalysisllms}, facts descriptions from criminal court opinions regarding thefts~\cite{drápal2023usinglargelanguagemodels}, and SMS health intervention prompts~\cite{prescott2024comparing}. Collaborative frameworks integrating human expertise with LLM capabilities have also been proposed to enhance the TA process~\cite{dai2023llm, zhang2023redefining, gao2024collabcoder}. Both open-source models (e.g., Llama 2, Mistral)~\cite{mathis2024inductive, katz2024thematic} and proprietary models (e.g., GPT-3.5, GPT-4, Claude)~\cite{de2024performing, dai2023llm, singh2024racer, mannstadt2024novel} have been utilized, demonstrating significant time savings and scalability while maintaining relevance~\cite{mathis2024inductive, prescott2024comparing}. Additionally, some studies have evaluated LLMs' capacity for deductive qualitative research~\cite{gao2024collabcoder, xiao2023supporting, singh2024racer}.

Despite significant advances, LLMs' ability to perform inductive TA on real-world healthcare interview transcripts remains under-explored. Prior studies have primarily focused on lower-stakes domains like video games or music~\cite{dai2023llm, de2024performing}. Even within healthcare, existing analyses often use small datasets with shorter transcripts, failing to capture the complexity of real-world scenarios~\cite{mathis2024inductive, mannstadt2024novel}. Moreover, evaluations of LLM-generated themes have not involved researchers who conducted the original inductive TA on the same dataset~\cite{mathis2024inductive, mannstadt2024novel}. Advanced LLMs and robust prompt engineering techniques have yet to be fully leveraged in this context~\cite{katz2024thematic, mathis2024inductive}. In this study, we introduce an \textbf{LLM}-Enhanced \textbf{T}hematic \textbf{A}nalysis (\textbf{LLM-TA}) pipeline and address these gaps with three key contributions:

\begin{itemize}
    \item We present the first LLM-enhanced TA pipeline tailored for high-stakes, lengthy, real-world, de-identified transcripts. Specifically, we analyze interview transcripts with parents of children diagnosed with Anomalous Aortic Origin of a Coronary Artery (AAOCA), a type of congenital heart disease.
    \item We apply thematic similarity analysis, LLM-based judgments, and assessments by a TA expert who previously worked on the dataset to compare LLM-generated themes with human-coded ground truth. Employing chunking strategies, we test various prompt engineering techniques—including zero-shot, one-shot, and reflection—on contextually rich AAOCA interview transcripts. Our LLM-TA pipeline outperforms existing LLM-augmented TA methods in thematic accuracy, LLM assessment, and expert review.
    \item  By closely collaborating with the inductive TA expert, we identified gaps of LLM-gernated themes compared to human researchers. We provide preliminary insights on integrating LLMs into real-world workflows to enhance the efficiency and scalability of inductive TA.
\end{itemize}

\section{Methodology}

We developed a \textbf{LLM}-Enhanced \textbf{T}hematic \textbf{A}nalysis pipeline (Figure \ref{fig:1}) to perform inductive TA~\cite{braun2006using} on de-identified transcripts from nine focus group sessions involving 42 parents. These transcripts, with a median word count of 11,457, document conversations between interviewers and parents of children with AAOCA. Traditional inductive TA involves six stages: \textit{1) Familiarization with the data, 2) Generation of initial codes, 3) Theme identification, 4) Theme review, 5) Final theme definition and naming, and, 6) Report production.} Stages \textit{(2)} through \textit{(5)} typically require at least two human researchers to independently generate and refine codes and themes, followed by collaborative discussions to finalize results. This iterative process is both time- and labor-intensive.

Formally, given a dataset of de-identified transcripts 
\[
T = \{ T_1, T_2, \dots, T_n \}
\]
where \( n = 9 \) focus group sessions, involving 42 parents of children with AAOCA, our objective is to automate the inductive TA process using an LLM. Each transcript \( T_i \) is lengthy (median word count of 11,457) and contains rich conversational data between parents. The goal is to generate a set of initial Codes and and Themes:
    \[
    C = \{ c_1, c_2, \dots, c_k \},
    \Theta = \{ \theta_1, \theta_2, \dots, \theta_k \},
    \]
that capture significant patterns in the data. There are two primary stages in the LLM-TA pipeline (Figure \ref{fig:1}):
\subsubsection{Stage 1: Initial code generation on chunked transcripts.}Codes are the foundational units of inductive TA, capturing significant concepts and ideas from the transcripts. Each code includes a concise name, a meaningful description, and representative quotes from the transcripts.
\begin{itemize}

\item \textbf{1(a) Splitting transcripts into smaller chunks.} To enable fine-grained analysis, we divided each transcript into smaller chunks of up to 1,500 words. Each chunk preserves the integrity of conversational context. This approach improves the LLM's ability to generate detailed and accurate codes, compared to directly processing entire transcripts as described by \citet{mathis2024inductive}.

\item \textbf{1(b) LLM-based initial coding.} We prompted the LLM to roleplay as an inductive TA researcher tasked with generating initial codes for each chunk. The prompt provided detailed context about the transcripts and instructed the LLM to identify exhaustive codes, each accompanied by a name, description, and representative quotes. On average, the LLM generated one code for every 225 words of transcript.

\item \textbf{1(c) Grouping initial codes.} After generating codes for all chunks, we concatenated the codes (including their names, descriptions, and quotes) and divided them into $N$ sequential groups. This divide-and-conquer strategy mitigates the LLM's limitations in processing large amounts of information at once~\cite{liu2023lostmiddlelanguagemodels,wang2024primacyeffectchatgpt}, ensuring that the grouped codes retain fine-grained details for subsequent theme generation.
\end{itemize}

\subsubsection{Stage 2: Theme Generation.} Themes represent overarching concepts synthesized from multiple related codes. Each theme includes a short title, a detailed description, and the associated codes. These themes are critical for understanding participants' perspectives and are used to inform the research findings.

\begin{itemize}
\item  \textbf{2(a) Preliminary theme generation.} For each group of codes, we prompted the LLM to synthesize preliminary themes. Acting as an inductive TA researcher, the LLM analyzed the grouped codes and their descriptions to identify themes. Each preliminary theme included a title, a detailed description, and the associated codes. This process was repeated for all $N$ groups, resulting in $N$ sets of preliminary themes. Notably, a single code could belong to multiple themes.
\item \textbf{2(b) Final theme generation.} In the final stage, the LLM reviewed all preliminary themes and their associated codes across groups. It synthesized overarching themes by considering the participants' experiences, needs, and meaningful outcomes related to living with children diagnosed with AAOCA. The final themes included detailed descriptions and were designed to reflect the most prominent insights from the data.
\end{itemize}

\subsubsection{Prompting Strategies.} To implement the pipeline, we employed three distinct prompting strategies of increasing complexity (Figure \ref{fig:2}): zero-shot, one-shot, and Reflexion~\cite{shinn2024reflexion}. All detailed prompts can be accessed through our public GitHub repository.

\begin{itemize}
    \item \textbf{Zero-shot prompting.} In the zero-shot setting, we used straightforward prompts to guide the LLM in identifying codes from transcript chunks and generating the final themes, as outlined in the pipeline. These prompts included a detailed explanation of the inductive TA methodology. Additionally, the prompts incorporated comprehensive background information on the AAOCA transcripts, aligning with the familiarization stage of traditional inductive TA. This approach allowed the LLM to process the data without requiring prior examples.

    \item \textbf{One-shot prompting.} Building on the zero-shot approach, the one-shot setting introduced curated examples from a related inductive TA study. These examples served as templates to guide the LLM in generating code names and theme descriptions that adhered to the desired format and context.

    \item \textbf{Reflexion prompting.} In the Reflexion setting~\cite{shinn2024reflexion}, we extended the one-shot approach by incorporating iterative feedback to refine the LLM's outputs. After the LLM generated preliminary themes, we prompted it to critically evaluate its outputs, focusing on the specificity and comprehensiveness of the theme names and descriptions. The LLM then generated feedback on its own outputs, identifying areas for improvement. Using this feedback, we conducted a second round of theme generation, refining the preliminary themes to ensure greater alignment with the the underlying data. A similar Reflexion-based strategy was employed during Stage 2(b) (Final theme generation) to synthesize final themes that were both coherent and representative of the participants' experiences.
\end{itemize}

\section{Experiment and Evaluation}

\begin{table*}[!ht]
    \centering
    \footnotesize
    \renewcommand{\arraystretch}{.8}
    \setlength{\tabcolsep}{5pt} 
    \begin{tabular}{llccc}
        \toprule
        \multicolumn{2}{l}{ } & \multicolumn{2}{c}{Human-labeled themes vs. LLM-labeled themes}\\
        & \textbf{Similarity Metrics} & \textbf{Jaccard Similarity} & \textbf{Hit Rate}\\ \midrule 
        
        \multirow{4}{*}{\citet{mathis2024inductive} (w/o Reflexion)} 
        & sentence-t5-xxl (\(>0.82\)) & 0.111& 0.667\\ 
        & all-mpnet-base-v2 (\(>0.62\)) & 0.146& 0.750\\ 
        & all-MiniLM-L6-v2 (\(>0.56\)) & 0.097& 0.500\\ 
        & LLM as a judge (\(>0.5\)) & 0.056& 0.667\\
 \midrule 
        
        \multirow{4}{*}{\citet{mathis2024inductive} (w/ Reflexion)} 
        & sentence-t5-xxl (\(>0.82\)) & 0.139& 0.583\\ 
        & all-mpnet-base-v2 (\(>0.62\)) & 0.139& 0.583\\ 
        & all-MiniLM-L6-v2 (\(>0.56\)) & 0.083& 0.417\\ 
        & LLM as a judge (\(>0.5\)) & 0.042& 0.750\\
 \midrule 
        
        \multirow{4}{*}{\textbf{LLM-TA (0-Shot)}} 
        & sentence-t5-xxl (\(>0.82\)) & \textbf{0.410}& \textbf{1.000}\\ 
        & all-mpnet-base-v2 (\(>0.62\)) & \textbf{0.410}& \textbf{0.917}\\ 
        & all-MiniLM-L6-v2 (\(>0.56\)) & \textbf{0.389}& \textbf{0.917}\\ 
        & LLM as a judge (\(>0.5\)) & 0.118& 0.750\\
 \midrule
        
        \multirow{4}{*}{\textbf{LLM-TA (1-Shot)}} 
        & sentence-t5-xxl (\(>0.82\)) & 0.396& \textbf{1.000}\\ 
        & all-mpnet-base-v2 (\(>0.62\)) & 0.326& \textbf{0.917}\\ 
        & all-MiniLM-L6-v2 (\(>0.56\)) & 0.285& \textbf{0.917}\\ 
        & LLM as a judge (\(>0.5\)) & \textbf{0.174}& \textbf{0.917}\\
 \midrule

        \multirow{4}{*}{\textbf{LLM-TA (1-Shot + Reflexion)}} 
        & sentence-t5-xxl (\(>0.82\)) & 0.222& \textbf{1.000}\\ 
        & all-mpnet-base-v2 (\(>0.62\)) & 0.291& 0.833\\ 
        & all-MiniLM-L6-v2 (\(>0.56\)) & 0.215& 0.833\\ 
        & LLM as a judge (\(>0.5\)) & 0.152& \textbf{0.917}\\
 \midrule
       
        \multirow{4}{*}{Ground Truth (Human)} 
        & sentence-t5-xxl (\(>0.82\)) & 0.583& 1.000\\ 
        & all-mpnet-base-v2 (\(>0.62\)) & 0.486& 1.000\\ 
        & all-MiniLM-L6-v2 (\(>0.56\)) & 0.500& 1.000\\ 
        & LLM as a judge (\(>0.5\)) & 0.104& 1.000\\ 
        \bottomrule
 
    \end{tabular}
    \caption{Performance comparison of human-generated and LLM-generated themes and descriptions. Jaccard similarity and Hit rate are used to measure Human-labeled themes vs. LLM-labeled themes based on pair-wise similarity in Figure \ref{fig:Continuous Heatmap}.}
    \label{tab:performance-comparison}
\end{table*}

\begin{table*}[!ht]
\footnotesize
    \renewcommand{\arraystretch}{.95}

    \centering
    \begin{tabular}{lcccc}
         \hline
         & \multicolumn{2}{c}{Similarity with Human Coding} & & \\ \cline{2-3}
         Methods & Concepts Level & Theme Level & Specificity & Usefulness \\ 
         \hline
         \citet{mathis2024inductive} (w/o Reflexion) & High & 1 (Low) & 1 (Low) & Not Very Helpful \\ 
         \citet{mathis2024inductive} (w/ Reflexion) & High & 1 (Low) & 2 (Medium) & Not Very Helpful \\ 
         \textbf{LLM-TA (0-Shot)}    & High & 2 (High) & 4 (High) & Helpful \\ 
         \textbf{LLM-TA (1-Shot)}    & High & 2 (High) & 3 (Moderate-High) & Moderately Helpful \\ 
         \textbf{LLM-TA (1-Shot + Reflexion)} & High & 3 (Highest) & 5 (Highest) & Helpful \\
         \hline
    \end{tabular}
    \caption{Expert evaluation.The LLM-TA pipeline outperformed baselines in thematic similarity, specificity, and usefulness.}
    \label{table:expert_eval}
\end{table*}

Qualitative research does not always have a definitive ground truth. In this study, to evaluate the performance of LLM-TA pipeline, we employ two complementary methods. First, we invited a core researcher, responsible for generating the human themes using the same dataset (see AAOCA patients focus group Transcript Dataset section), to provide qualitative feedback regarding the accuracy and helpfulness of the LLM-generated themes and descriptions. We also utilize embedding similarity and LLM judgment methods to evaluate the similarity between the LLM-generated and human-generated themes and theme descriptions.

\begin{figure*}[!ht]
    \centering
    \includegraphics[width=.9\linewidth]{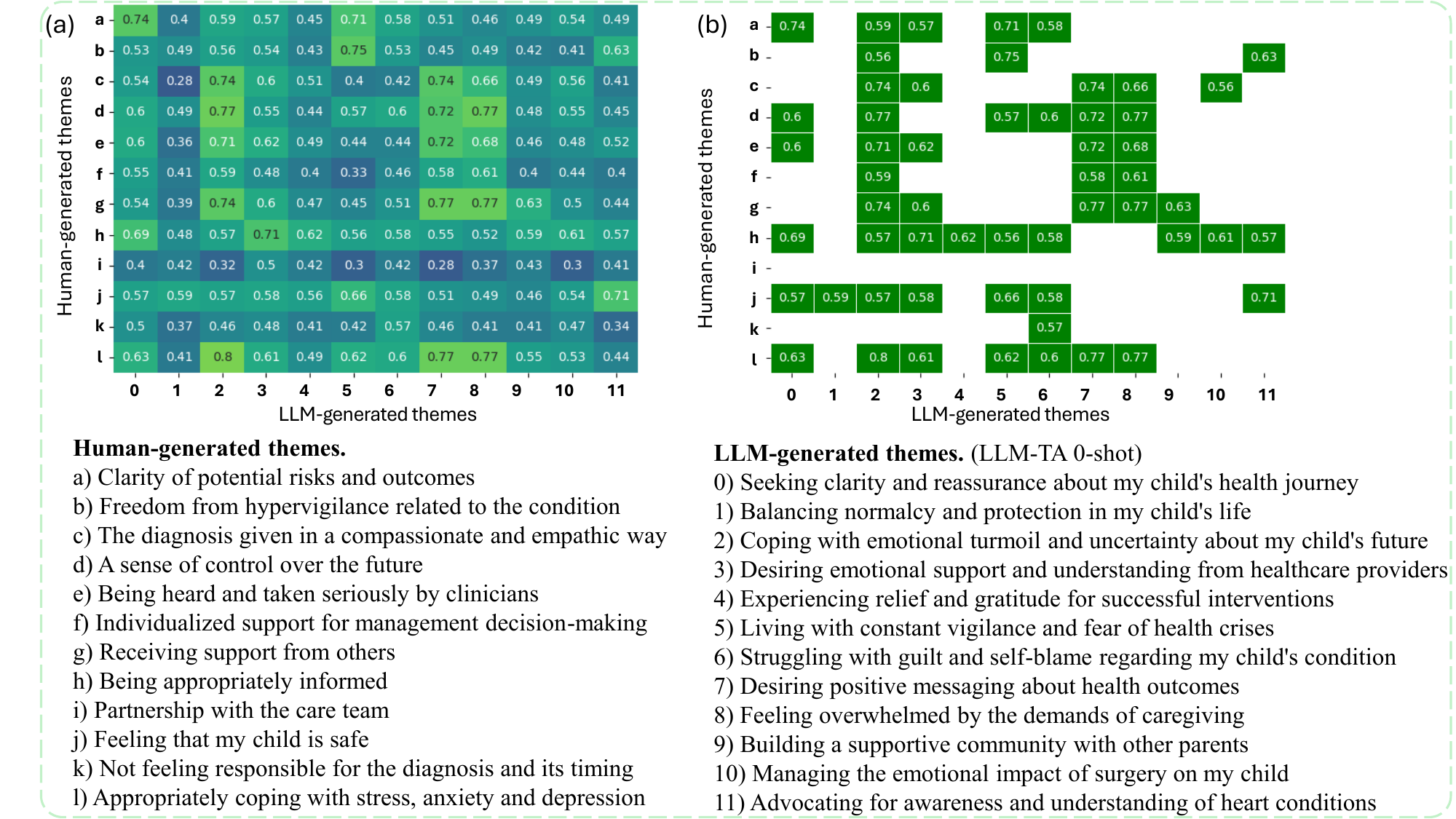}
    \caption{(a) Similarity heatmap of themes generated by the LLM-TA (zero-shot) method compared with human themes using all-MiniLM-L6-v2. (b) Similarity heatmap of themes generated by the LLM-TA (zero-shot) method compared with human themes using all-MiniLM-L6-v2, with a threshold score of 0.56.}
    \label{fig:Continuous Heatmap}
\end{figure*}
\textbf{Similarity Metrics.} We employed both embedding and LLM-based metrics to validate the LLM-TA pipeline by evaluating the similarity between human and LLM-generated descriptions. For each method, pairwise similarity scores between theme descriptions were aggregated into similarity matrices. Following \citet{mathis2024inductive}, we converted continuous similarity scores into binary classifications (similar or not) by setting a threshold through sensitivity analysis (Figure \ref{fig:Continuous Heatmap}). Based on the similarity matrices, we calculated the \emph{Jaccard Similarity} and the \emph{Hit Rate}.

Let \( H = \{ h_1, h_2, \dots, h_n \} \) represent the set of human-generated themes, and \( L = \{ l_1, l_2, \dots, l_m \} \) represent the set of LLM-generated themes. For each pair \( (h_i, l_j) \) in \( H \times L \), we compute a similarity score \( s(h_i, l_j) \). We define \( S_\theta = \{ (h_i, l_j) \in H \times L \mid s(h_i, l_j) \geq \theta \} \), where \( \theta \) is the similarity threshold determined via sensitivity analysis and kept consistent across different baselines.

The \emph{Jaccard Similarity} is defined as the proportion of theme pairs considered similar out of all possible pairs:
\begin{equation}
\text{Jaccard Similarity} = \frac{ \left| S_\theta \right| }{ \left| H \times L \right| } = \frac{ \left| S_\theta \right| }{ n \times m } \tag{1}
\end{equation}

The \emph{Hit Rate} measures the proportion of human-generated themes that find a highly similar mapping in the LLM-generated themes:
\begin{equation}
\text{Hit Rate} = \frac{ \left| H_s \right| }{ n } \tag{2}
\end{equation}
where \( H_s = \{ h \in H \mid \exists\, l \in L,\, s(h, l) \geq \theta \} \).

These metrics provide insights into the similarity between human-generated themes \( H \) and LLM-generated themes \( L \). The Jaccard Similarity quantifies the proportion of similar theme pairs while the Hit Rate indicates how many human themes are adequately captured by the LLM.





\textbf{Embedding-Based Semantic Similarity.} We employed the sentence transformer models: \texttt{all-MiniLM-L6-v2}, \texttt{all-mpnet-base-v2}, and \texttt{sentence-T5-xxl}, to encode the theme descriptions into high-dimensional embeddings. Pairwise cosine similarity scores were computed between human-generated and LLM-generated embeddings.

\textbf{LLM-Based Similarity.} The LLM acts as a judge to assess the similarity of ideas between human-generated and LLM-generated themes~\cite{10.1145/3678884.3681850}. For each pair of themes, the LLM assigned a similarity score between 0 and 1 based on its understanding of the conceptual overlap, with a score of 0 meaning completely different and 1 meaning completely overlapping. We prompted the LLM to apply a penalty if one description is very specific and the other is very general, ensuring a more balanced evaluation.

\textbf{Baselines.} We used the method proposed by \citet{mathis2024inductive} as our baseline, which incorporates iterative refinement, chain-of-thought, and Reflexion prompting. We also performed a simplified version of their method without the Reflexion module, while keeping all other settings identical.

\textbf{Experimental Settings.} For all our proposed methods, we use OpenAI's \texttt{GPT-4o-mini-2024-07-18} with a temperature of 0 for reproducibility. We build the pipeline using \texttt{langchain v0.3.21}. For the baseline methods proposed by \citet{mathis2024inductive}, we follow their strategy and set the temperature to 1.0.

\subsubsection{AAOCA Patients Focus Group Transcript Dataset.} The de-identified transcript corpus was derived from nine focus group sessions involving 42 parents of children with AAOCA. These sessions were lightly moderated by a non-clinical facilitator, encouraging open discussions that allowed parents to express previously unarticulated needs and experiences related to living with the condition. The nine transcripts had a median word count of 11,457. Using traditional inductive TA methods, three study team members independently coded the transcripts and developed themes based on the data. They engaged in iterative discussions to review and refine codes and identify themes. This process was both time- and resource-intensive and required approximately 30 person-hours. The final analysis identified twelve meaningful outcomes for individuals and parents.

For this study, we used these nine deidentified transcripts as our dataset. Additionally, one of the experts from the original inductive TA team collaborated closely with us to evaluate the LLM-generated results. The twelve themes representing meaningful outcomes, as articulated by the participants and identified by the human research team, serve as the ground truth data. The theme titles are listed in Figure \ref{fig:Continuous Heatmap}.

\section{Results}
\textbf{Efficiency Analysis}. Conducting inductive TA on nine lengthy transcripts required approximately 30 person-hours for human researchers. In contrast, the LLM-TA pipeline completed the task in under 10 minutes for all methods except the 1-shot + Reflexion approach, which took 90 minutes (1.5 hours). Even this most time-intensive method reduced task duration by 97\% compared to manual analysis. While some quality gaps remain between LLM- and human-generated themes, the results highlight the potential for integrating the LLM-TA method into human-led inductive TA workflows to enhance scalability and efficiency.

\textbf{Quantitative Evaluation}. \emph{Jaccard Similarity} and \emph{hit-rate} were used for evaluation. Table~\ref{tab:performance-comparison} summarizes the comparison between human- and LLM-generated themes, utilizing five prompting and chunking strategies. Two baseline methods were replicated from the~\cite{mathis2024inductive} study, alongside three variants of our proposed LLM-TA pipeline. Our approaches showed clear improvements across both metrics. For \emph{Jaccard Similarity}, the zero-shot strategy achieved the highest score in three of four evaluation methods, as well as the highest overall average. For \emph{hit-rate}, the one-shot method performed the best, obtaining the highest or joint-highest scores across all evaluation methods. Performance declined slightly in the one-shot + Reflexion setting. Compared to baseline methods, our pipeline achieved a ~216\% improvement in average \emph{Jaccard Similarity} and ~45\% improvement in average \emph{hit-rate}.

Two key factors underpinned the superior performance of our approach. First, the chunking strategy played a pivotal role. Baseline methods~\cite{mathis2024inductive} analyzed shorter transcripts with a median length of ~3,200 words and did not utilize chunking. While this approach is adequate for smaller datasets, it struggles with large, real-world transcripts. By dividing longer transcripts into manageable chunks, our method prevented information loss and facilitated the extraction of more nuanced insights, thereby improving theme generation. Second, the use of detailed prompts significantly enhanced outcomes. Unlike the generic prompts used in baseline methods, our prompts explicitly outlined the research context, objectives, and task-specific instructions, such as generating theme names, descriptions, and identifying relevant quotes. This precision enabled the LLM to produce outputs more aligned with human-generated themes. Moreover, in the one-shot setting, we further enriched contextual understanding by incorporating a real-world example from Inductive TA conducted on similar transcripts. This example clarified expectations and reinforced the LLM's ability to structure outputs effectively.


\textbf{Expert Evaluation}. To assess the LLM-generated themes, we engaged a human researcher who had previously conducted the ground truth inductive TA on the same AAOCA dataset. This expert's domain familiarity enabled a precise evaluation of thematic accuracy and relevance.

Table~\ref{table:expert_eval} highlights the expert’s overall evaluation. The LLM-TA pipeline outperformed baselines in thematic similarity, specificity, and usefulness. However, the expert noted limitations in the utility of themes generated by the one-shot LLM-TA method, despite its superior hit rate. In contrast, themes produced by the zero-shot setting were qualitatively more useful, underscoring the insufficiency of similarity-based metrics for assessing thematic quality.


The expert also identified some shortcomings in LLM-generated themes. \textbf{(1) Lack of representativeness.} Certain themes overemphasized rare, dramatic experiences rather than reflecting broader trends. For instance, the theme \textit{“Finding relief in a diagnosis after health crises”} disproportionately highlighted cardiac arrest cases, whereas most parents of incidentally diagnosed children reported heightened anxiety instead of relief. Similarly, themes like \textit{“Managing the financial challenges of care”} overrepresented financial concerns, which were rarely mentioned across transcripts. \textbf{(2) Inaccurate interpretations.} Some themes misrepresented transcript content. For example, the theme \textit{“Seeking simplicity in discussing my child’s heart condition”} misinterpreted parents’ preferences for a comprehensive understanding as a desire for simplified information. Similarly, the overly broad theme \textit{“Desiring clear communication and understanding from medical professionals”} conflated distinct issues such as empathy and clinical communication into a single category. \textbf{(3) Missing clinical context.} The LLM lacked knowledge of clinical context external to the transcripts, such as long-term outcomes for AAOCA or management strategy nuances. For example, parents’ desire for more information reflects a lack of clinical evidence on optimal management strategies of AAOCA, rather than mere gaps in communication. Those clinical contexts are not in the transcripts, and providing the model with such clinical context could improve the accuracy of future LLM-generated themes.

These findings underscore the necessity of close collaboration with domain experts to iteratively refine prompts based on research goals and contextual knowledge. High-stakes datasets often require expertise-derived context beyond what transcripts explicitly provide, which LLMs inherently lack. Our results reinforce the irreplaceable role of expert human evaluation in TA, particularly in complex healthcare datasets. While embedding-based and LLM-based similarity metrics provide valuable quantitative insights, only human reviewers can, as of now, qualitatively assess thematic relevance and accuracy.

\section{Conclusion}
Motivated by the need to reduce the time-consuming inductive TA process in complex healthcare transcript data, we designed an LLM-enhanced inductive TA pipeline to automate stages (1) to (5) of the traditional TA process. Using chunking, we performed fine-grained coding and theme identification. Additionally, we developed detailed instructions on the steps of TA and the research goals, specifically regarding the meaningful outcomes for parents of children with AAOCA, and incorporated these into various LLM prompting strategies. By evaluating the pipeline with domain expertise and LLM-based metrics, we significantly improved the performance of existing LLM-enhanced TA methods. We emphasize that the evaluation of LLM-enhanced inductive TA on high-stakes data must involve domain experts. While automated methods can assist in analyzing correlations and provide quantitative insights, expert evaluation remains irreplaceable and invaluable in ensuring the accuracy and relevance of themes derived from such datasets.

Through close collaboration with an inductive TA expert, we identified key limitations of current LLM-generated themes, primarily arising from the lack of clinical context that domain experts possess and cannot be directly inferred from transcripts. Based on these findings, we strongly recommend that future researchers collaborate closely with domain experts to design prompts and interpret results at each stage of TA. To further enhance reliability, future work should: (1) engage multiple experts to mitigate individual biases and broaden interpretive perspectives, (2) incorporate clinical guidelines or regulatory documents into prompts to enrich LLMs’ contextual understanding, and (3) validate the pipeline across diverse medical conditions to assess generalizability. Such steps will help fully harness LLMs’ potential to scale inductive TA while preserving analytical rigor in sensitive healthcare domains.

\subsection{Acknowledgements}
We would like to acknowledge the following funding supports: NIH OT2OD032581, NIH OTA-21-008, NIH 1OT2OD032742-01.

\bibliography{aaai25}

\end{document}